\documentclass[10pt,twocolumn,letterpaper]{article}

\usepackage[pagenumbers]{cvpr} 
\usepackage{adjustbox} 
\usepackage{multirow}
\usepackage{colortbl}      
\usepackage{xcolor}
\usepackage{tikz}

\usepackage{booktabs}      
\usepackage{multirow}      
\usepackage{array}         
\usepackage{graphicx}      
\usepackage{adjustbox}     

\usepackage[table]{xcolor} 
\usepackage{colortbl}      

\usepackage{makecell}      

\usepackage{caption}
\usepackage{float}

\definecolor{myred}{RGB}{255,102,102}
\definecolor{myyellow}{RGB}{255,230,128}
\definecolor{mygreen}{RGB}{153,255,153}
\definecolor{myblue}{RGB}{153,204,255}
\definecolor{myorange}{RGB}{255,178,102}
\definecolor{mygray}{gray}{0.9}

\definecolor{cvprblue}{rgb}{0.21,0.49,0.74}
\usepackage[pagebackref,breaklinks,colorlinks,allcolors=cvprblue]{hyperref}

\def\paperID{} 
\def\confName{CVPR}
\def\confYear{2026}

\title{ Unified Personalized Understanding, Generating and Editing}

\author{
  \begin{tabular}{>{\centering}p{0.9\textwidth}}
  Yu Zhong\textsuperscript{1},
  Tianwei Lin\textsuperscript{1},
  Ruike Zhu\textsuperscript{1},
  Yuqian Yuan\textsuperscript{1},
  Haoyu Zheng\textsuperscript{1},
  Liang Liang\textsuperscript{1},
  Wenqiao Zhang\textsuperscript{1,${\dag}$},
  Feifei Shao\textsuperscript{1},
  Haoyuan Li\textsuperscript{2},
  Wanggui He\textsuperscript{2},
  Hao Jiang\textsuperscript{2},
  Yueting Zhuang\textsuperscript{1}
  \end{tabular} \\ \\[-1pt]
  \begin{tabular}{>{\centering}p{0.9\textwidth}}
  \textsuperscript{1}Zhejiang University,
  \textsuperscript{2}Alibaba Group
  \end{tabular} \\[6pt]
}

\makeatletter
\let\@oldmaketitle\@maketitle

\renewcommand{\@maketitle}{\@oldmaketitle
\vspace{-10mm}
\centering
\includegraphics[width=1\linewidth]{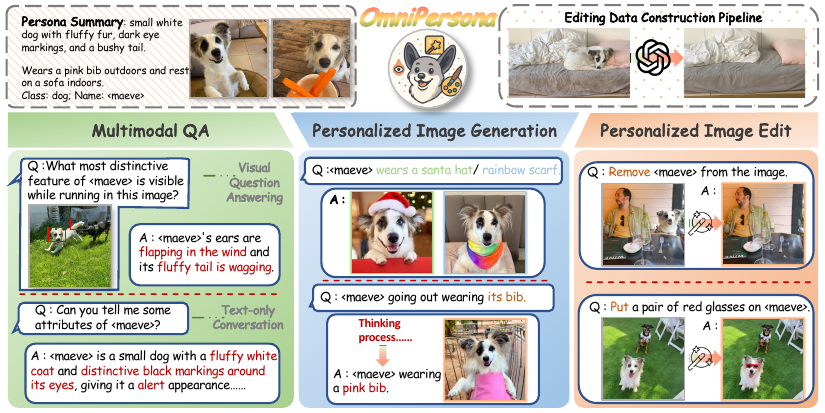}
\vspace{-7mm}
\captionof{figure}{\textbf{Capability Overview of OmniPersona.} OmniPersona leverages decoupled learnable prompts to achieve \textbf{unified personalized understanding, generation, and editing} from only a few concept images paired with textual descriptions. Notably, OmniPersona is the first framework to enable personalized image editing, a critical capability overlooked by previous works.
}
\vspace{3mm}

\label{fig:overview}
}
\makeatletter

\makeatletter
\def\blfootnote{\gdef\@thefnmark{}\@footnotetext}
\makeatother

\begin{document}
\maketitle
\blfootnote{${\dag}$ denotes corresponding author}

\begin{abstract}
\vspace{-8mm}

Unified large multimodal models (LMMs) have achieved remarkable progress in general-purpose multimodal understanding and generation. However, they still operate under a ``one-size-fits-all'' paradigm and struggle to model user-specific concepts (e.g., generate a photo of \texttt{<maeve>}) in a consistent and controllable manner. Existing personalization methods typically rely on external retrieval, which is inefficient and poorly integrated into unified multimodal pipelines. Recent personalized unified models introduce learnable soft prompts to encode concept information, yet they either couple understanding and generation or depend on complex multi-stage training, leading to cross-task interference and ultimately to fuzzy or misaligned personalized knowledge.
We present \textbf{OmniPersona}, an end-to-end personalization framework for unified LMMs that, for the first time, integrates personalized understanding, generation, and image editing within a single architecture. OmniPersona introduces structurally decoupled concept tokens, allocating dedicated subspaces for different tasks to minimize interference, and incorporates an explicit knowledge replay mechanism that propagates personalized attribute knowledge across tasks, enabling consistent personalized behavior.
To systematically evaluate unified personalization, we propose \textbf{\texttt{OmniPBench}}, extending the public UnifyBench concept set with personalized editing tasks and cross-task evaluation protocols integrating understanding, generation, and editing. Experimental results demonstrate that OmniPersona delivers competitive and robust performance across diverse personalization tasks. We hope OmniPersona will serve as a strong baseline and spur further research on controllable, unified personalization.
\end{abstract}    
\section{Introduction}
\label{sec:Introduction}
Recently, unified large multimodal models~(LMMs) for joint understanding and generation have demonstrated remarkable potential as general-purpose AI assistants~\cite{zhang2025unified,lu2024unified,ge2024seed,zhou2024transfusion,tong2025metamorph}. Models such as Chameleon~\cite{team2024chameleon}, Janus-Series~\cite{wu2024janus,ma2024janusflow,chen2025janus}, Blip3-o~\cite{chen2025blip3}, Show-o2~\cite{xie2025show} and Bagel~\cite{deng2025emerging} can reliably follow user instructions across diverse task formats, indicating that unified LMMs exhibit strong generalization ability over a broad spectrum of multimodal tasks in open-world settings. In realistic human--AI interaction, however, user demands are often organized around their own entities and preferences, requiring assistants to understand and persistently memorize personalized concepts and to align with user intent under brief and naturals instructions. Existing unified models remain substantially constrained in such user-level scenarios. For example, when using the identifier \texttt{<maeve>} for a specific dog, even with attribute descriptions, current models still struggle to handle \texttt{<maeve>} consistently and controllably. Consequently, robustly integrating personalized concepts into unified LMMs is crucial for building truly personal AI assistants.

To enable personalization in LMMs, some efforts have been made. 
One line of work uses retrieval-augmented generation~(RAG)~\cite{hao2025rap}, injecting concept attributes and descriptions as external context; another learns soft prompts to encode specific concepts in the latent space of the model~\cite{alaluf2024myvlm,nguyen2024yo}.

In parallel, several methods propose dedicated personalization mechanisms for unified LMMs and report promising results on certain benchmarks~\cite{an2025unictokens,nguyen2025yo}, yet three key challenges remain for achieving comprehensive and robust personalization in unified modeling:

\noindent\textbf{(i) Coupled and Conflicting Representations.} In unified LMMs, sharing a common parameter space across tasks naturally induces representational conflict~\cite{wu2024janus,lin2025healthgpt,zhang2025unified}, which is further exacerbated in few-shot personalization, where a single concept representation must simultaneously support understanding, generation, and editing. Existing methods (e.g., UniToken~\cite{jiao2025unitoken}) mitigate such interference via multi-stage or alternating training, but personalized information remains compressed into a single latent space, without structurally distinguishable ``slots'' to host task-specific solution spaces. As a result, it is difficult to selectively leverage pretrained knowledge for understanding versus generation in a targeted and decoupled manner.

\begin{figure}[t]

\centering
\includegraphics[width=1\linewidth]{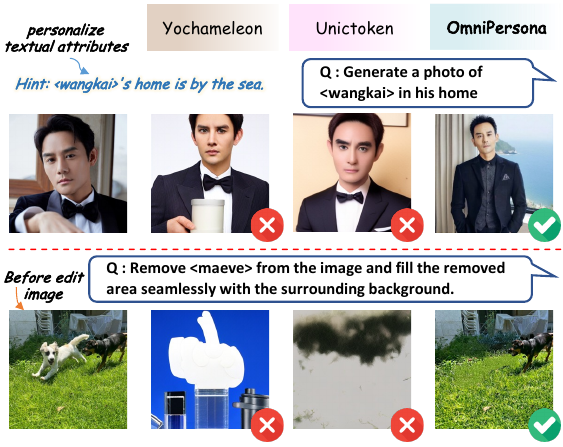}
\vspace{-7mm}
\captionof{figure}{Current state-of-the-art personalized unified models fail to stably leverage textual knowledge during generation, resulting in images misaligned with concept descriptions (top). Moreover, existing methods neglect personalized image editing, producing outputs unrelated to the input image (bottom).
}
\vspace{-4mm}

\label{fig:case_intro}
\end{figure}

\noindent\textbf{(ii) Opaque Knowledge Latents}. Existing methods compress concept knowledge into opaque embeddings that lack explicit interpretability---we cannot verify what knowledge is encoded or how it drives the model's understanding and generation. This opacity poses a critical challenge for \textit{personalized attribute-reasoning generation (PARG)}~\cite{an2025unictokens}, where models generate images that must leverage learned textual attributes (e.g., ``\texttt{<wangkai>} in his home'' without specifying the home's features). As shown in Fig.~\ref{fig:case_intro}, even approaches like Yo'Chameleon~\cite{nguyen2025yo} and UniCTokens~\cite{an2025unictokens}, which attempt to enhance knowledge expression via implicit prompts and shared tokens, struggle to distinguish whether the model genuinely leverages textual attributes or merely memorizes visual patterns from training data. Developing interpretable knowledge representations that enable genuine PARG capabilities remains an open challenge.

\noindent\textbf{(iii) Personalized Editing Gap}. Prior works do not examine the feasibility of personalized image editing~(see Fig.~\ref{fig:case_intro}). However, editing is arguably the most demanding test of personalized tasks: the model must follow user instructions to perform local or structural modifications while accurately locating and preserving the target identity~\cite{zheng2024makima, pan2025wiseedit}, effectively coupling understanding and generation within a single operation. At the same time, current benchmarks neither systematically evaluate this crucial capability nor investigate whether incorporating editing data can in turn enhance performance across other personalization tasks.

To address these challenges, we propose \textbf{OmniPersona}, a novel end-to-end framework for unified personalized understanding, generation, and editing. The proposed framework is composed of the following key components:

\textbf{Token Representations Decoupling.} 
We relax the assumption that a single shared parameter space suffices for personalization by introducing structurally decoupled concept tokens and assigning them to task-specialized expert subspaces, so that the same personalized concept is captured by related but distinct tokens along different task pathways. This decoupling design mitigates cross-task interference and enables selective reuse of pretrained knowledge for understanding versus generation.

\textbf{Explicit Knowledge Replay.} 
OmniPersona introduces a reasoning-based knowledge replay mechanism that externalizes personalized knowledge as explicit intermediate representations. Specifically, the model self-generates descriptions of concept attributes as conditioning signals and replays the concept--text--image pathway, explicitly aligning knowledge acquired in understanding with downstream generation and enforcing semantic consistency of the concept across unified tasks.

\textbf{Understanding--Generation--Editing Synergy.} 
We emphasize personalized editing as a crucial component and investigate whether the fine-grained constraints imposed by editing can in turn regularize and strengthen unified personalized representations. To this end, we propose \textbf{\texttt{OmniPBench}}, which extends the existing unified personalization benchmark UnifyBench~\cite{an2025unictokens} with carefully curated personalized editing data and cross-task evaluation protocols spanning understanding, generation, and editing, enabling a systematic assessment of how editing supervision benefits overall personalization performance.

In summary, our contributions are as follows:
\begin{itemize}
  \item To the best of our knowledge, the proposed \textbf{OmniPersona} serves as the first end-to-end unified framework that simultaneously achieves personalized image understanding, generation, and editing.
  \item Our framework combines decoupled token representations to reduce task conflict, and uses inference-time explicit knowledge replay to improve personalized attribute-reasoning generation stability and interpretability.
  \item We propose \textbf{\texttt{OmniPBench}} with integrated editing data to systematically assess synergy across tasks.
  \item Comprehensive experiments demonstrate competitive performance across personalized tasks and state-of-the-art (SOTA) results in personalized editing.
\end{itemize}

\section{Related Work}
\label{sec:Related work}

\textbf{Unified Large Multimodal Models.}
As multimodal understanding and generation advance, unified models become increasingly crucial across domains~\cite{zhang2022boostmis, lv2023duet, zhang2024revisiting}. In recent years, unified large multimodal models have attracted significant attention~\cite{lu2024unified,wang2024emu3,wu2024vila,ma2025unitok,wu2025qwen, ma2024janusflow}. Early works such as Janus-Pro~\cite{chen2025janus} and UniToken~\cite{jiao2025unitoken} adopt a pure autoregressive paradigm, converting multiple modalities into discrete vocabularies and performing joint modeling through a shared decoding process, thereby unifying multimodal inputs and outputs within a common semantic space. However, these approaches often suffer from low generation efficiency and limited flexibility in content synthesis. To enhance generation quality and training efficiency, several studies such as SEED-X~\cite{ge2024seed}, MetaQueries~\cite{pan2025transfer}, and BLIP3-o~\cite{chen2025blip3} introduce diffusion-based generation heads, which enable rapid alignment between understanding and generation capabilities using limited multimodal paired data, and even demonstrate synergistic improvements across tasks. Nevertheless, these methods remain constrained by the sequential decoding bottleneck, making real-time high-resolution generation challenging. To address this issue, subsequent works (e.g., Transfusion~\cite{zhou2024transfusion}, Show-o~\cite{xie2025show} and BAGEL~\cite{deng2025emerging}) combine the strengths of autoregressive and diffusion models to construct hybrid architectures that maintain the unified multimodal framework while balancing generation flexibility and computational efficiency. As unified models continue to mature, an emerging research direction lies in effectively customizing and adapting them to recognize and leverage user-specific concepts, enabling personalized understanding and generation within a unified modeling paradigm.

\noindent\textbf{Personalized Multimodal Understanding and Generation.} 
 Personalized adaptation aims to enable models to recognize and utilize specific concepts, achieving rapid user-level semantic alignment~\cite{yuan2025videorefer, yuan2025eoc, yuan2025pixelrefer, zhang2024hyperllava, zhang2021consensus}. MyVLM~\cite{alaluf2024myvlm} and Yo'LLaVA~\cite{nguyen2024yo} inject personalized concepts through learnable soft prompts, avoiding lengthy contextual descriptions while enhancing understanding capability. Retrieval-augmented approaches such as LaMP~\cite{salemi2024lamp} and RAP~\cite{hao2025rap} treat conceptual profiles as external memory, leveraging contextual retrieval to improve personalized reasoning. MC-LLaVA~\cite{an2024mc} further optimizes multi-concept discrimination and co-occurrence understanding through joint training and visual token initialization, enhancing generalization in multi-entity scenarios. On the generation side, works such as DreamBooth~\cite{ruiz2023dreambooth} achieve high-fidelity subject-driven image synthesis via few-shot fine-tuning of text-to-image diffusion models, laying the foundation for personalized visual generation~\cite{shi2024instantbooth,he2024imagine,ye2023ip}. More recently, Yo'Chameleon~\cite{nguyen2025yo} unifies personalized understanding and generation within a single framework through dual soft prompts and a self-prompting mechanism, while UniCTokens~\cite{an2025unictokens} integrates complementary semantics of both tasks via unified concept tokens and a three-stage mutual learning paradigm, achieving higher efficiency and performance in unified personalized modeling. However, existing approaches still lack an end-to-end personalized modeling mechanism that can fully exploit the potential of unified multimodal architectures.

\vspace{-2mm}

\section{Methodology}
\label{sec:Method}

We propose an end-to-end personalized multimodal framework that unifies understanding, generation, and editing with token representations decoupling, explicit knowledge replay, and understanding--generation--editing synergy.

\definecolor{sksblue}{RGB}{65, 105, 225}      
\definecolor{undgreen}{RGB}{46, 139, 87}      
\definecolor{genpurple}{RGB}{138, 43, 226}    

\subsection{Token Representations Decoupling}
\label{sec:token-routing}

To mitigate the representational entanglement discussed in Sec.~\ref{sec:Introduction}, we introduce token representations decoupling. Instead of learning a unified set of prompts, we route task-specific learnable tokens to specialized expert subspaces, enabling each branch to develop modality-aware representations while preserving cross-task synergy underpinned by the inherent modality-interaction of the unified model.

\begin{figure}[t]
\centering
\includegraphics[width=1\linewidth]{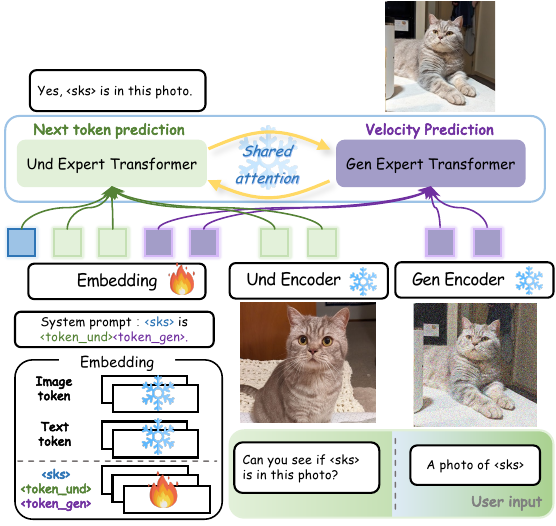}
\vspace{-7mm}
\captionof{figure}{\textbf{Model Overview of OmniPersona.} We employ an end-to-end dual-branch routing mechanism to train two sets of parameter-decoupled prompts.
}
\vspace{-3mm}
\label{fig:method}
\end{figure}

Formally, during fine-tuning, we represent each personalized concept using a special identifier (e.g., \texttt{<sks>}) composed of multiple learnable tokens that are routed to different expert branches. The system prompt is structured as:
\[
\resizebox{0.9\linewidth}{!}{``$\textcolor{sksblue}{\texttt{<sks>}} \texttt{is}\, \textcolor{undgreen}{\texttt{<und\_1>}} \cdots \textcolor{undgreen}{\texttt{<und\_}N_u\texttt{>}} \; \textcolor{genpurple}{\texttt{<gen\_1>}} \cdots \textcolor{genpurple}{\texttt{<gen\_}N_g\texttt{>}}$.''}
\]
where tokens enclosed in angle brackets are learnable concept embeddings. Specifically, as shown in Fig.~\ref{fig:overview}, the concept identifier \texttt{<sks>} together with \texttt{<und\_i>} tokens are routed to the \textit{understanding expert}, while \texttt{<gen\_j>} tokens are routed to the \textit{generation expert}. Let
\begin{align*}
\mathbf{P}^{(\text{und})} &= [\mathbf{p}_{\text{sks}}, \mathbf{p}_1^{(\text{und})}, \dots, \mathbf{p}_{N_u}^{(\text{und})}] \in \mathbb{R}^{(N_u+1) \times d}, \\
\mathbf{P}^{(\text{gen})} &= [\mathbf{p}_1^{(\text{gen})}, \dots, \mathbf{p}_{N_g}^{(\text{gen})}] \in \mathbb{R}^{N_g \times d},
\end{align*}
denote the embedding matrices for the two expert-specific token subsets, where $d$ is the embedding dimension. Note that $\mathbf{p}_{\text{sks}}$ is included in the understanding expert to enable concept identification.
During the forward pass, each expert processes only its routed prompt subset and corresponding input embeddings:
\begin{equation}
\resizebox{0.9\linewidth}{!}{$
\mathbf{H}^{(\text{und})} =
\mathcal{F}_{\text{und}}\!\left(\mathbf{P}^{(\text{und})}, \mathbf{X}^{(\text{und})}\right), \quad
\mathbf{H}^{(\text{gen})} =
\mathcal{F}_{\text{gen}}\!\left(\mathbf{P}^{(\text{gen})}, \mathbf{X}^{(\text{gen})}\right),
$}
\end{equation}
where \(\mathcal{F}_{\text{und}}(\cdot)\) and \(\mathcal{F}_{\text{gen}}(\cdot)\) denote the expert branches of the unified model for understanding and generation, respectively, with $\mathbf{X}^{(\text{und})}$ and $\mathbf{X}^{(\text{gen})}$ representing the corresponding personalized instructions.

This routing formulation enforces expert-level parameter decoupling, enabling each branch to specialize on its designated modality while preserving synergy under a universal unified model architecture.

\subsection{Explicit Knowledge Replay}
\label{sec:replay}

\begin{figure}[tb]

\centering
\includegraphics[width=1\linewidth]{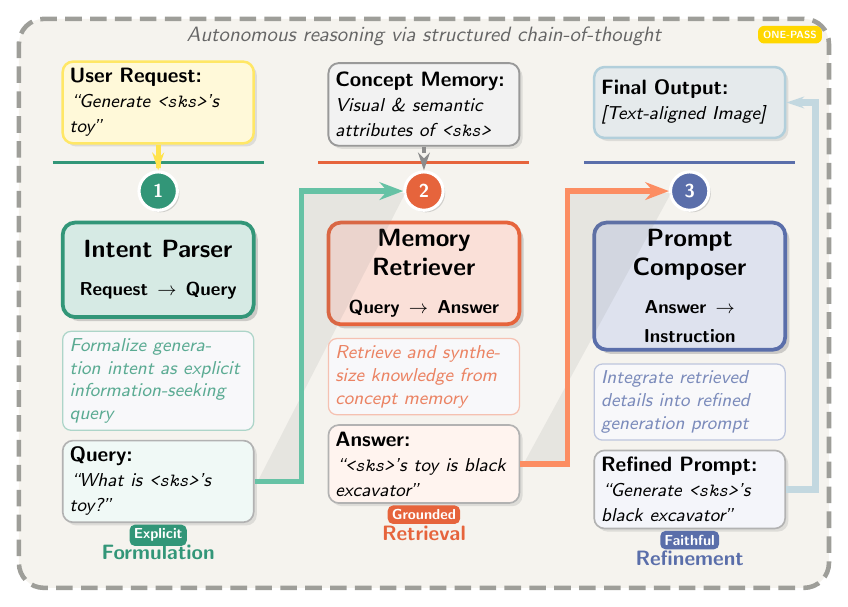}
\vspace{-7mm}
\captionof{figure}{Explicit knowledge replay pipeline.
}
\vspace{-3mm}
\label{fig:agent}
\end{figure}

While token decoupling mitigates cross-task interference, the learned embeddings remain opaque---it is unclear whether the model genuinely leverages semantic knowledge or merely memorizes visual patterns. To address this, we propose explicit knowledge replay, an \textit{inference-time} mechanism that actively retrieves and externalizes concept knowledge. As shown in Fig.~\ref{fig:agent}, we introduce a three-stage framework that transforms implicit generation requests into explicit, grounded, and faithful prompts. 
Importantly, all the aforementioned stages are recursively executed within a single unified multimodal model $\mathcal{UMM}$, achieving one-step generation.

\noindent\textbf{Stage 1: Intent Parser.}
Given the original user request $T$, we reformulate it into an explicit information-seeking query using $\mathcal{UMM}$ in text mode:
\begin{equation}
\mathcal{Q} = \mathcal{UMM}^{\text{text}}(T),
\end{equation}
which transforms generation intent into explicit queries $\mathcal{Q}$ such as ``What is \texttt{<sks>}'s [attribute]?''.

\noindent\textbf{Stage 2: Memory Retriever.}
The formulated query $\mathcal{Q}$ is used to retrieve concept memory from the learned token representations $\mathbf{P}$:
\begin{equation}
\mathcal{A} = \mathcal{UMM}^{\text{text}}(\mathcal{Q}, \mathbf{P}),
\end{equation}
where $\mathcal{A}$ represents the grounded answer, e.g., ``\texttt{<sks>}'s toy is a black excavator.''

\noindent\textbf{Stage 3: Prompt Composer.}
The retrieved answer $\mathcal{A}$ is integrated back into the generation context:
\begin{equation}
\hat{T} = \mathcal{UMM}^{\text{text}}(\mathcal{A}, T),
\end{equation}
producing a refined, faithful prompt such as ``Generate \texttt{<sks>}' black excavator.''

\noindent\textbf{Final Generation.}
The refined prompt $\hat{T}$ is passed through the model's generation mode for personalized synthesis:
\begin{equation}
I_{\text{gen}} = \mathcal{UMM}^{\text{gen}}(\hat{T}, \mathbf{P}).
\end{equation}

This unified pipeline ensures outputs are (1) \textit{explicit} through query formulation, (2) \textit{grounded} through memory retrieval, (3) \textit{faithful} through refinement, and (4) \textit{personalized} through concept-aware generation, all within a single model in one pass.

\subsection{Understanding-Generation-Editing Synergy}

While existing personalized multimodal models focus on understanding and text-to-image generation, they overlook the editing dimension and lack suitable datasets for personalized editing.
To enhance conceptual understanding and editing capability, we introduce a \textbf{personalized image editing dataset}.

We construct a personalized editing dataset $\mathcal{D}_{\text{edit}}$, where each sample is a triplet $(I_{\text{src}}, E_{\text{edit}}, I_{\text{tgt}})$ consisting of a source image $I_{\text{src}}$, an editing instruction $E_{\text{edit}}$ (e.g., "remove \texttt{<sks>} in the photo"), and the target image $I_{\text{tgt}}$.
The model is trained end-to-end, jointly optimizing all soft tokens across understanding, generation, and editing tasks.

\textbf{Training Objectives.}
The model is jointly optimized via two complementary losses:

\noindent\textbf{(i) Cross-Entropy Loss} for autoregressive token prediction based on the personalized understanding task:
\begin{equation}
\mathcal{L}_{\text{text}}^{\text{CE}} = - \sum_{i=1}^{C} x_i \log \hat{x}_i,
\end{equation}
where $x_i$ and $\hat{x}_i$ denote the ground-truth and predicted token distributions, and $C$ is the number of classes.

\noindent\textbf{(ii) Mean Squared Error Loss} following the Rectified Flow paradigm. Given a clean latent $\mathbf{x}_0$ and noise $\mathbf{x}_1$, we obtain the interpolated latent $\mathbf{x}_t = (1 - t)\mathbf{x}_0 + t\mathbf{x}_1$ for $t \in [0, 1]$. The MSE objective for image generation is:
\begin{equation}
\mathcal{L}_{\text{image}}^{\text{MSE}} =
\mathbb{E}\!\left[\, \lVert g_{\theta}(\mathbf{x}_t \mid c) - (\mathbf{x}_0 - \mathbf{x}_1) \rVert_2^2 \right],
\end{equation}
where $g_{\theta}$ predicts the velocity field conditioned on $\mathbf{x}_t$ and context $c$.
Similarly, the editing loss is also defined as MSE:
\begin{equation}
\mathcal{L}_{\text{edit}}^{\text{MSE}} =
\mathbb{E}\!\left[\, \lVert g_{\theta}(\mathbf{x}_t \mid c_{\text{edit}}) - (\mathbf{x}_0^{\text{tgt}} - \mathbf{x}_1^{\text{tgt}}) \rVert_2^2 \right],
\end{equation}
where $c_{\text{edit}}$ includes the source image and editing instruction, and $\mathbf{x}_0^{\text{tgt}}$ is the target latent.

\textbf{Joint Training with Editing Data.}
The total objective combines understanding, generation, editing optimizations:
\begin{equation}
\mathcal{L}_{\text{total}} =
\mathcal{L}_{\text{text}}^{\text{CE}} + \lambda_{\text{image}} \, \mathcal{L}_{\text{image}}^{\text{MSE}} + \lambda_{\text{edit}} \, \mathcal{L}_{\text{edit}}^{\text{MSE}},
\end{equation}
where $\lambda_{\text{image}}$ and $\lambda_{\text{edit}}$ control the relative contribution of image generation and editing.

\begin{table*}[!t]
\caption{Quantitative comparison on \textbf{\texttt{OmniPBench}}. TP = Text Prompt. IP = Image Prompt. PARG = Personalized Attribute-Reasoning Generation. SEMA-C = Semantic Consistency. QUAL-I = Quality of Image. We compare open-source unified models; others are for \colorbox{gray!20}{reference}. The \colorbox{red!15}{best} are highlighted.}
\label{tab:unifiedbench}
\centering
\setlength{\tabcolsep}{0.4mm}
\renewcommand{\arraystretch}{1.25}
\begin{adjustbox}{width=\textwidth, center}
\begin{tabular}{l|c|c|
c|cccc|
ccc|c|
cc|
ccc}
\Xhline{2\arrayrulewidth}
\multirow{4}{*}{\textbf{Method}} &
\multirow{4}{*}{\textbf{Token}} &
\multirow{4}{*}{\begin{tabular}[c]{@{}c@{}}\textbf{Training}\\ \textbf{Images}\end{tabular}} &
\multicolumn{5}{c|}{\textbf{Personalized Understanding}} &
\multicolumn{4}{c|}{\textbf{Personalized Generation}} &
\multicolumn{2}{c|}{\multirow{3}{*}{\textbf{PARG}}} &
\multicolumn{3}{c}{\multirow{3}{*}{\textbf{Personalized Edit}}} \\
\cmidrule(lr){4-8}\cmidrule(lr){9-12}
 &  &  &
\multicolumn{1}{c|}{\textbf{Rec.}} & \multicolumn{2}{c}{\textbf{VQA}} & \multicolumn{2}{c|}{\textbf{QA}} &
\multicolumn{3}{c|}{\textbf{Pure Gen.}} & \textbf{People Gen.} &
 &  &
 &  &  \\
\cmidrule(lr){4-17}
 &  &  &
Weight & BLEU & GPT & BLEU & GPT &
CLIP-I & CLIP-T & DINO & Face-Simi &
Score & CLIP-I &
SEMA-C & QUAL-I & Avg. \\[-2pt]
\hline
\rowcolor{red!8}
\multicolumn{17}{c}{\textbf{\textit{Upper Bound}}}\\
\hline
\textcolor{gray}{GPT-4o+TP} & \textcolor{gray}{$\sim$100} & \textcolor{gray}{-} &
\textcolor{gray}{0.742} & \textcolor{gray}{0.473} & \textcolor{gray}{0.676} & \textcolor{gray}{0.610} & \textcolor{gray}{0.685} &
\textcolor{gray}{0.689} & \textcolor{gray}{0.301} & \textcolor{gray}{0.626} & \textcolor{gray}{0.198} &
\textcolor{gray}{0.780} & \textcolor{gray}{0.690} & \textcolor{gray}{0.463} & \textcolor{gray}{0.645} & \textcolor{gray}{0.554} \\
\textcolor{gray}{GPT-4o+IP} & \textcolor{gray}{$\sim$1{,}000} & \textcolor{gray}{-} &
\textcolor{gray}{0.773} & \textcolor{gray}{0.543} & \textcolor{gray}{0.685} & \textcolor{gray}{0.589} & \textcolor{gray}{0.652} &
\textcolor{gray}{0.794} & \textcolor{gray}{0.310} & \textcolor{gray}{0.722} & \textcolor{gray}{0.559} &
\textcolor{gray}{0.782} & \textcolor{gray}{0.792} & \textcolor{gray}{0.512} & \textcolor{gray}{0.603} & \textcolor{gray}{0.558} \\
Real Images & - & - &
- & - & - & - & - &
0.832 & - & 0.728 & 0.739 &
- & 0.832 & - & - & - \\
\Xhline{1\arrayrulewidth}

\rowcolor{orange!15}
\multicolumn{17}{c}{\textbf{\textit{Understanding Only}}}\\
\hline
\textcolor{gray}{Yo'LLaVA} & \textcolor{gray}{16} & \textcolor{gray}{$\sim$100} &
\textcolor{gray}{0.919} & \textcolor{gray}{0.609} & \textcolor{gray}{0.629} & \textcolor{gray}{0.612} & \textcolor{gray}{0.593} &
- & - & - & - & - & - & - & - & - \\
\textcolor{gray}{MC-LLaVA} & \textcolor{gray}{16} & \textcolor{gray}{$\sim$10} &
\textcolor{gray}{0.924} & \textcolor{gray}{0.628} & \textcolor{gray}{0.637} & \textcolor{gray}{0.601} & \textcolor{gray}{0.583} &
- & - & - & - & - & - & - & - & - \\
\textcolor{gray}{RAP-MLLM} & \textcolor{gray}{$\sim$1{,}000} & \textcolor{gray}{-} &
\textcolor{gray}{0.940} & \textcolor{gray}{0.616} & \textcolor{gray}{0.616} & \textcolor{gray}{0.712} & \textcolor{gray}{0.722} &
- & - & - & - & - & - & - & - & - \\
\textcolor{gray}{Qwen2.5-VL + TP} & \textcolor{gray}{$\sim$100} & \textcolor{gray}{-} &
\textcolor{gray}{0.660} & \textcolor{gray}{0.407} & \textcolor{gray}{0.727} & \textcolor{gray}{0.574} & \textcolor{gray}{0.774} &
- & - & - & - & - & - & - & - & - \\
\textcolor{gray}{Yo'LLaVA(Phi-1.5)} & \textcolor{gray}{16} & \textcolor{gray}{$\sim$100} &
\textcolor{gray}{0.765} & \textcolor{gray}{0.488} & \textcolor{gray}{0.497} & \textcolor{gray}{0.510} & \textcolor{gray}{0.494} &
- & - & - & - & - & - & - & - & - \\
\Xhline{1\arrayrulewidth}

\rowcolor{green!12}
\multicolumn{17}{c}{\textbf{\textit{Generation Only}}}\\
\hline
\textcolor{gray}{Text inversion} & \textcolor{gray}{-} & \textcolor{gray}{$\sim$10} &
- & - & - & - & - &
\textcolor{gray}{0.630} & \textcolor{gray}{0.247} & \textcolor{gray}{0.569} & \textcolor{gray}{0.371} &
\textcolor{gray}{0.070} & \textcolor{gray}{0.628} & - & - & - \\
\textcolor{gray}{DreamBooth (SD)} & \textcolor{gray}{-} & \textcolor{gray}{$\sim$10} &
- & - & - & - & - &
\textcolor{gray}{0.649} & \textcolor{gray}{0.281} & \textcolor{gray}{0.591} & \textcolor{gray}{0.436} &
\textcolor{gray}{0.071} & \textcolor{gray}{0.650} & - & - & - \\
\Xhline{1\arrayrulewidth}

\rowcolor{cyan!8}
\multicolumn{17}{c}{\textbf{\textit{Unified Model}}}\\
\hline
Chameleon+TP & $\sim$100 & - &
0.690 & 0.413 & 0.488 & 0.509 & 0.564 &
0.547 & 0.176 & 0.509 & 0.011 &
0.329 & 0.549 & 0.122 & 0.520 & 0.321 \\
Chameleon+IP & $\sim$1{,}000 & - &
0.493 & 0.445 & 0.498 & 0.407 & 0.535 &
0.523 & 0.160 & 0.469 & 0.066 &
0.299 & 0.499 & 0.093 & 0.442 & 0.268 \\
Show-o+TP & $\sim$100 & - &
0.566 & 0.461 & 0.409 & 0.504 & 0.579 &
0.664 & 0.264 & 0.553 & 0.048 &
0.770 & 0.660 & 0.250 & 0.340 & 0.295 \\
Bagel+TP & $\sim$100 &  - & 0.788 & 0.420 & 0.542 & 0.445 & 0.632 & 0.697 & \colorbox{red!15}{0.284} & 0.425 & 0.309 & \colorbox{red!15}{0.813} & 0.725 & 0.297 & 0.566 & 0.432\\
Yo'Chameleon & 32 & $\sim$1{,}000 &
0.764 & 0.474 & 0.507 & 0.510 & 0.581 &
0.697 & 0.236 & 0.590 & 0.224 &
0.266 & 0.698 & 0.108 & 0.360 & 0.234 \\
Unictoken & 32 & $\sim$10 &
0.790 & 0.503 & 0.523 & 0.544 & 0.603 &
0.750 & 0.282 & \colorbox{red!15}{0.646} & 0.334 &
0.359 & 0.749 & 0.155 & 0.212 & 0.184 \\

\textbf{OmniPersona (Ours)} & \textbf{32} & \textbf{$\sim$10} &
\colorbox{red!15}{0.852} & \colorbox{red!15}{0.529} & \colorbox{red!15}{0.603} & \colorbox{red!15}{0.646} & \colorbox{red!15}{0.678} &
\colorbox{red!15}{0.791} & {0.273} & 0.579 & \colorbox{red!15}{0.413} &
{0.613} & \colorbox{red!15}{0.788} & \colorbox{red!15}{0.711} & \colorbox{red!15}{0.605} & \colorbox{red!15}{0.658} \\
\Xhline{2\arrayrulewidth}
\end{tabular}
\end{adjustbox}
\vspace{-3mm}
\end{table*}

\section{Experiment}
\label{sec:Experiment}

We conduct comprehensive experiments to evaluate our framework across four personalized tasks: \textit{understanding}, \textit{generation}, \textit{personalized attribute-reasoning generation}, and \textit{image editing}, where editing represents a novel evaluation dimension absent in prior benchmarks.

\noindent\textbf{Implementation Details.}
We allocate $N=32$ learnable tokens per concept, split equally between understanding-specific tokens ($N_u=16$) and generation-specific tokens ($N_g=16$). We optimize using AdamW for 2{,}000 steps per concept with batch size 8. All experiments are conducted on H20 GPUs using Bagel~\cite{deng2025emerging} as the backbone model. Additional training details are provided in the Appendix.

\noindent\textbf{Dataset.}
We introduce \textbf{\texttt{OmniPBench}}, a comprehensive benchmark for unified personalization spanning understanding, generation, and editing. Building upon the 20 concepts from UnifyBench~\cite{an2025unictokens}, which include Person (10), Pets (5), and Objects (5), we contribute a novel \textbf{personalized editing dataset}. For each concept with identifier \texttt{<sks>} and class label $c$, we automatically generate removal instructions following the template ``remove \texttt{<sks>:} $c$ from the photo''. Target images are synthesized via inpainting models. We perform manual verification, with 90\% of generated pairs passing quality control. Moreover, we design diverse edit-instruction templates to systematically probe each concept's editing robustness under various modification scenarios. Additional details are provided in the Appendix.

\noindent\textbf{Baselines.}
We compare against four categories of personalization methods: \textit{understanding-only} approaches~\cite{alaluf2024myvlm,nguyen2024yo}, \textit{generation-only} methods~\cite{ruiz2023dreambooth,gal2022image}, \textit{unified personalization} frameworks~\cite{an2025unictokens,nguyen2025yo}, and \textit{retrieval-augmented} systems~\cite{hao2025rap}. To establish context-aware performance, we additionally include \textit{Bagel} with text-only descriptions (no learned tokens) as a zero-shot baseline.

\noindent\textbf{Metrics.}
For personalized image editing, our key contribution, we employ an LLM-as-judge protocol to assess semantic alignment between edited images and textual instructions, and edit quality evaluated by visual naturalness. For understanding, generation, and personalized attribute-reasoning generation, we adopt standard UnifyBench metrics. All results are reported as mean scores across concepts.

\subsection{Personalized Understanding and Generation}

As shown in Table~\ref{tab:unifiedbench}, our method achieves strong performance across understanding and generation tasks. Compared to SOTA unified models, we outperform Unictoken by 7.8\% on recognition and 13.1\% on average across question-answering tasks (VQA and text-based QA). For generation, we achieve the highest CLIP-I score (0.791), indicating superior alignment with personalized visual concepts, alongside substantial improvements in identity-critical metrics: face similarity (0.413 vs. 0.334).

While our DINO (0.579) is lower than Unictoken (0.646) and our CLIP-T (0.273) is lower than Bagel+TP (0.284), this reflects our design priority: \textit{preserving concept identity over generic text alignment}. The highest CLIP-I validates this trade-off, and our CLIP-T still outperforms other unified models (Show-o: 0.264, Yo'Chameleon: 0.236), confirming that identity-first optimization does not compromise overall text-image consistency. Compared to Bagel+TP (0.788), which directly injects concept descriptions as long context, we achieve superior understanding performance (0.852) with more compact token representations, demonstrating more efficient knowledge internalization.

\begin{figure*}[t]
\centering
\includegraphics[width=1\linewidth]{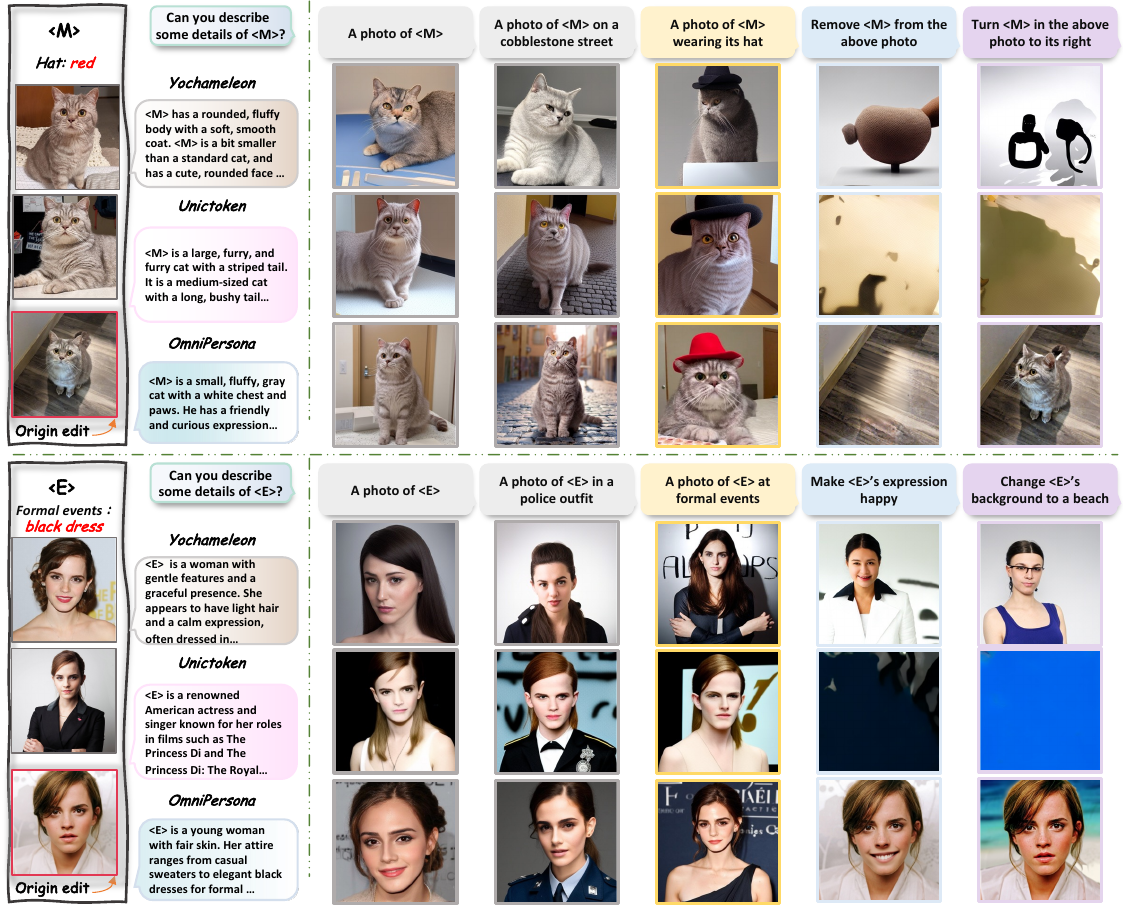}
\vspace{-5mm}
\captionof{figure}{Qualitative comparison with Yo'Chameleon and Unictoken on \textbf{\texttt{OmniPBench}}. \textbf{OmniPersona (Ours)} demonstrates more precise, personalized image generation and excellent capability of personalized image editing.
}
\vspace{-3mm}
\label{fig:feature}
\end{figure*}

\subsection{Personalized Attribute-Reasoning Generation}

Table~\ref{tab:unifiedbench} evaluates personalized attribute-reasoning generation~(PARG), where the model must leverage concept-related textual knowledge during generation. For example, given ``Generate a photo of \texttt{<wangkai>} in his home", the model must recall that \texttt{<wangkai>}'s home is by the sea, a piece of knowledge that is absent from visual priors. This tests explicit textual knowledge integration rather than image memorization.

With our explicit knowledge replay mechanism, we achieve SOTA CLIP-I (0.788), outperforming Unictoken (0.749) by 5.2\%, demonstrating superior identity preservation in knowledge-conditioned generation. While Bagel+TP achieves a higher overall PARG score (0.813 vs. 0.613) by directly injecting long-context descriptions, our approach internalizes knowledge in compact learnable tokens without increasing inference overhead. Notably, our score (0.613) substantially surpasses other learned-token methods like Unictoken (0.359) by 70.8\%, validating that explicit knowledge replay effectively bridges understanding and generation.

\subsection{Personalized Image Editing}

We introduce the first systematic evaluation of personalized image editing for unified models. As shown in Table~\ref{tab:unifiedbench}, we assess editing capability through LLM-as-judge metrics measuring (1) semantic alignment with instructions (SEMA-C) and (2) edited image quality (QUAL-I).

Our method achieves SOTA overall editing performance (0.658), surpassing GPT-4o+IP by 17.9\% (0.558) and demonstrating the effectiveness of our unified architecture for identity-preserving editing. Notably, we excel in semantic alignment (SEMA-C: 0.711), substantially outperforming all baselines including Bagel+TP (0.297), validating that our decoupled token representations enable precise instruction following while maintaining concept fidelity. Our competitive image quality (QUAL-I: 0.605) further confirms that explicit knowledge replay preserves visual coherence during complex editing operations.

\section{Ablation Studies and Analysis}
\label{sec:Ablation Studies}

\begin{table*}[!t]
\caption{Ablation study on \textbf{\texttt{OmniPBench}}.}
\label{tab:ablation}
\centering
\setlength{\tabcolsep}{0.4mm}
\renewcommand{\arraystretch}{1.25}
\begin{adjustbox}{width=\textwidth, center}
\begin{tabular}{l|
c|cccc|
ccc|c|
cc|
ccc}
\Xhline{2\arrayrulewidth}
\multirow{4}{*}{\textbf{Methods}} &
\multicolumn{5}{c|}{\textbf{Personalized Understanding}} &
\multicolumn{4}{c|}{\textbf{Personalized Generation}} &
\multicolumn{2}{c|}{\multirow{3}{*}{\textbf{PARG}}} &
\multicolumn{3}{c}{\multirow{3}{*}{\textbf{Personalized Edit}}} \\
\cmidrule(lr){2-6}\cmidrule(lr){7-10}
 &
\multicolumn{1}{c|}{\textbf{Rec.}} & \multicolumn{2}{c}{\textbf{VQA}} & \multicolumn{2}{c|}{\textbf{QA}} &
\multicolumn{3}{c|}{\textbf{Pure Gen.}} & \textbf{People Gen.} &
 &  &
 &  &  \\
\cmidrule(lr){2-15}
 &
Weight & BLEU & GPT & BLEU & GPT &
CLIP-I & CLIP-T & DINO & Face-Simi &
Score & CLIP-I &
SEMA-C & Qual-I & Avg. \\[-2pt]

\Xhline{1\arrayrulewidth}
\rowcolor{gray!10}
W/o Disentangled Tokens & 0.804 & 0.477 & 0.541 & 0.589 & 0.652 &
0.778 & 0.289 & 0.585 & 0.402 &
0.431 & 0.732 & 0.652 & 0.613 & 0.633 \\

W/o Knowledge Replay & - & - & - & - & - &
- & - & - & - &
0.312 & 0.765 & - & - & - \\

\rowcolor{gray!10}
W/o Edit Data & 0.840 & 0.517 & 0.614 & 0.632 & 0.659 &
0.772 & 0.296 & 0.553 & 0.397 &
0.584 & 0.781 & 0.683 & 0.592 & 0.638 \\

Ours (Full) & 0.852 & 0.529 & 0.603 & 0.646 & 0.678 &
0.791 & 0.273 & 0.579 & 0.413 &
0.613 & 0.788 & 0.711 & 0.605 & 0.658 \\

\Xhline{2\arrayrulewidth}
\end{tabular}
\end{adjustbox}
\vspace{-2mm}
\end{table*}

\vspace{2mm}
\begin{figure*}
  \centering
  \begin{subfigure}{0.68\linewidth}
    \includegraphics[width=1\linewidth]{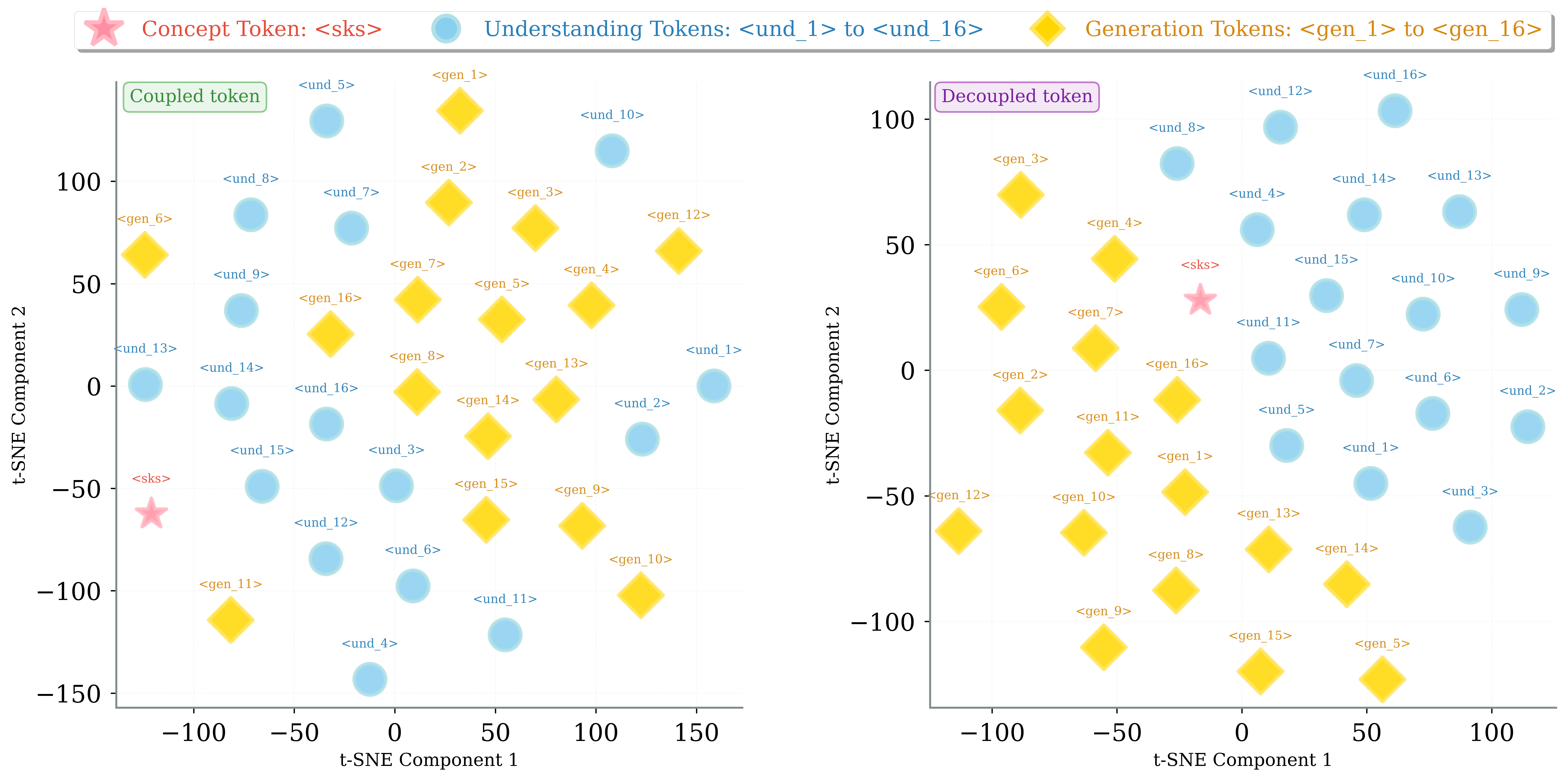}
    \caption{}
    \label{fig:short-a}
  \end{subfigure}
  \hfill
  \begin{subfigure}{0.28\linewidth}
  \includegraphics[width=1\linewidth]{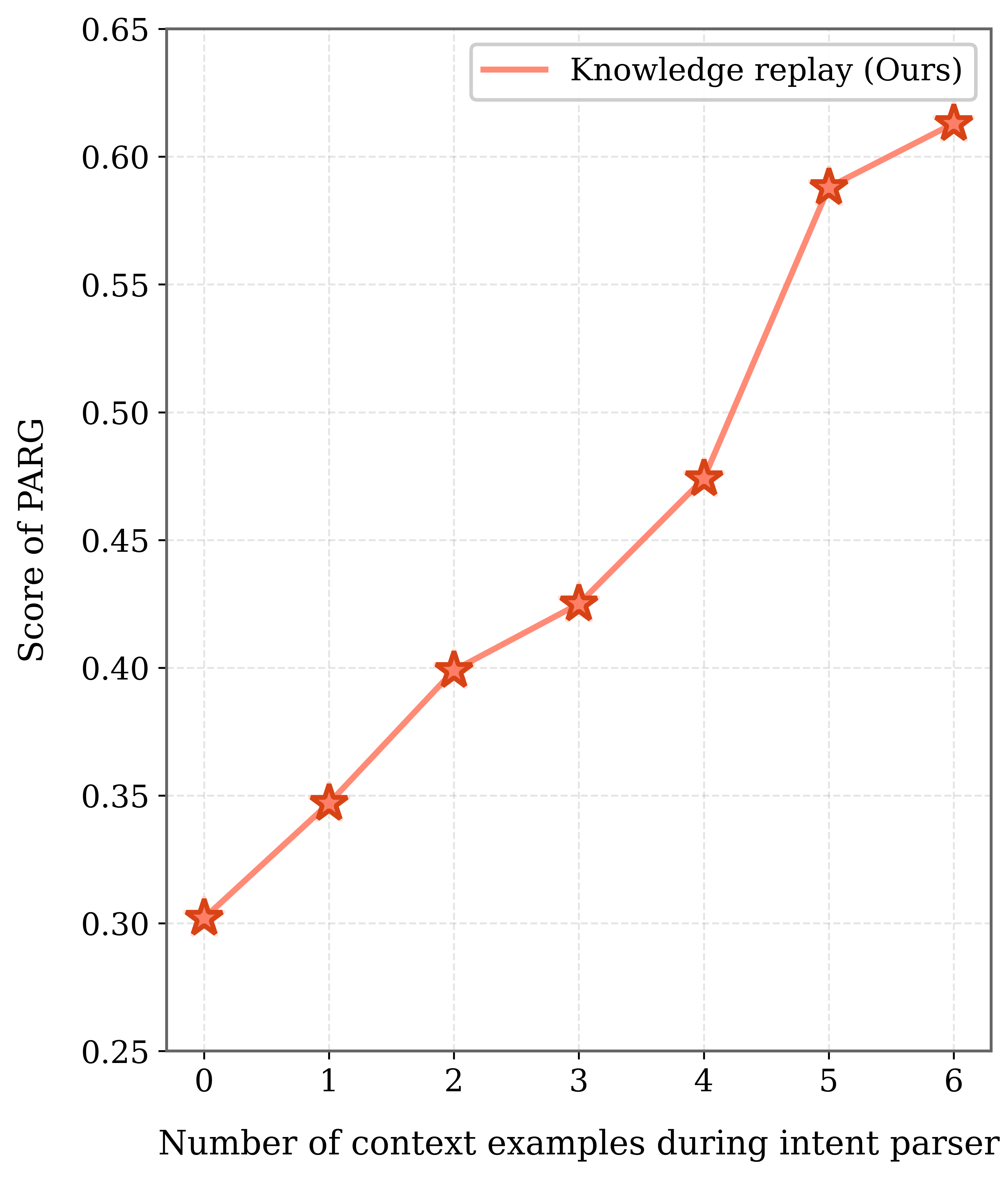}
    \caption{}
    \label{fig:short-b}
  \end{subfigure}
  \vspace{-3mm}
  \caption{Token embedding visualization via t-SNE (a) and the impact of in-context exemplar quantity on PARG score (b). 
  }
  \vspace{-5mm}
  \label{fig:quality-ablation}
\end{figure*}

\textbf{Decoupling Feature Space.}
Fig.~\ref{fig:quality-ablation} and Table~\ref{tab:ablation} validate our token decoupling mechanism. Without disentangled tokens, all embeddings share a unified transformer, causing severe representational entanglement (Fig.~\ref{fig:quality-ablation} (a), left), where overlapping clusters indicate destructive task interference. This directly degrades understanding performance: recognition drops 5.6\% (0.852$\to$0.804), VQA BLEU decreases 9.8\% (0.529$\to$0.477), and VQA GPT falls 10.3\% (0.603$\to$0.541). Understanding tasks suffer disproportionately because concept features become contaminated by generation-specific gradients in the shared latent space, forcing conflicting optimization objectives.

Our decoupled architecture (Fig.~\ref{fig:quality-ablation} (a), right) produces well-separated clusters, routing understanding and generation tokens to task-specialized expert subspaces. This enables selective reuse of pretrained knowledge: understanding tokens leverage semantic reasoning without interference from synthesis priors. The consistent gains across all understanding metrics validate our core insight: structurally distinguishable token ``slots" mitigate gradient conflicts and unlock the model's capacity for unified personalization.

\textbf{Knowledge Replay Promotes Personalized Knowledge Flow.}
Table~\ref{tab:ablation} validates the critical role of knowledge replay: without it, PARG score plummets 49.1\% (0.613$\to$0.312) while CLIP-I remains relatively stable (0.788$\to$0.765). This reveals that implicit embeddings preserve identity but fail to leverage semantic knowledge, demonstrating that explicit text externalization is essential for personalized attribute-reasoning generation.

Fig.~\ref{fig:quality-ablation} (b) analyzes the Intent Parser's generalization through in-context learning. As exemplars increase from 0 to 6, generation score rises consistently from 0.30 to 0.62 (+106.7\%), demonstrating that structured demonstrations enable the parser to accurately reformulate diverse user intents into explicit knowledge queries. This validates that in-context learning strengthens the concept$\to$text$\to$image pathway, enabling robust knowledge-grounded synthesis regardless of input phrasing.

\textbf{Multi-task Collaboration Yields Cross-task Gains.}
During training, we observe that models trained solely on understanding and generation spontaneously acquire preliminary editing capabilities, which motivates us to investigate whether editing supervision could reciprocally enhance understanding and generation. Table~\ref{tab:ablation} confirms this hypothesis: incorporating editing data improves recognition by 1.4\% (0.840$\to$0.852), QA GPT by 2.9\% (0.659$\to$0.678), and face similarity by 4.0\% (0.397$\to$0.413), alongside editing performance gains of 3.1\% (0.638$\to$0.658).

This bidirectional synergy stems from editing's unique constraint structure, which requires simultaneous concept \textit{localization} (understanding), identity \textit{preservation} (generation), and instruction \textit{alignment}. Unlike isolated tasks where the model can rely on visual pattern memorization, editing forces concept tokens to encode disentangled representations capturing both invariant identity attributes and modifiable contextual properties. These fine-grained constraints regularize identity preservation under distribution shift, transferring to all personalized generation scenarios. Additional details are provided in the Appendix.
\section{Conclusion}
\label{sec:Conclusion}
We present \textbf{OmniPersona}, the first end-to-end framework unifying personalized understanding, generation, and image editing within unified multimodal models. Our framework introduces token representations decoupling to mitigate cross-task interference, explicit knowledge replay to enable personalized attribute-reasoning generation, and demonstrates understanding--generation--editing synergy through joint training strategies.
We propose \textbf{\texttt{OmniPBench}}, the first benchmark for unified personalization evaluation, extending UnifyBench with personalized editing tasks and cross-task protocols. Experiments demonstrate competitive performance in personalized tasks.
This work represents a significant step toward truly personal AI assistants capable of consistent, controllable concept modeling.

\newpage

{
    \small
    \bibliographystyle{ieeenat_fullname}
    \bibliography{main}
}


\clearpage
\setcounter{page}{1}
\maketitlesupplementary

\appendix

\section{Details of Dataset}
\label{sec:dataset_details}

We manually constructed five types of editing operations, each accounting for 1/5 of the evaluation set, which are: object manipulation (e.g., Remove \texttt{<sks>}), attribute modification (e.g., Make \texttt{<sks>} raise their hand.), spatial transformation (e.g., Show \texttt{<sks>} from the side view.), environment interaction (e.g., Make \texttt{<sks>} appear in sunset lighting.), style appearance (e.g., Make \texttt{<sks>} look like a pencil sketch.)

\begin{figure}[htbp]

\centering
\includegraphics[width=0.8\linewidth]{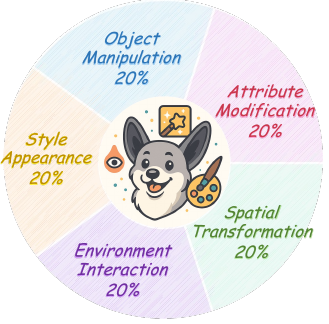}
\vspace{-3mm}
\captionof{figure}{Distribution of five editing operation types in the OmniPBench evaluation set, each accounting for 20\% of the dataset.
}

\vspace{-3mm}
\label{fig:edit_eval}
\end{figure}

\section{Details of Explicit Knowledge Replay}
\label{sec:knowledge_replay_details}
To maintain consistency in expression, we need to clarify that the knowledge-driven generation mentioned at the beginning of Section~\ref{sec:Experiment} of the main paper refers to Personalized Attribute-Reasoning Generation.
Then, we provide a comprehensive analysis of our explicit knowledge replay mechanism (Section~\ref{sec:replay}), addressing inference efficiency and prompt design.

\noindent\textbf{Computational Overhead Analysis.}
Section~\ref{sec:replay} describes a conceptual three-stage pipeline (Intent Parser $\to$ Memory Retriever $\to$ Prompt Composer). Actually we implement this as a \textbf{single unified prompt} to minimize latency. Table~\ref{tab:replay_overhead} compares sequential versus unified implementations.
Our unified single-stage implementation requires only \textbf{1.45s}, achieving a \textbf{1.29$\times$ speedup} over the sequential approach (1.87s) while providing substantial quality gains (+96.5\% PARG score over baseline).

\begin{table}[h]
\centering
\resizebox{1\linewidth}{!}{
\begin{tabular}{lccc}
\toprule
Implementation & Latency (s) & Model Calls & PARG Score \\
\midrule
\textit{Sequential Three-Stage} & & & \\
\quad Intent Parser & 0.54 & 1 & - \\
\quad Memory Retriever & 0.49 & 1 & - \\
\quad Prompt Composer & 0.84 & 1 & - \\
\quad Total & 1.87 & \textbf{3} & \textbf{0.635} \\
\midrule
\textit{Unified Single-Stage (Ours)} & \textbf{1.45} & \textbf{1} & 0.613 \\
\midrule
\textit{Baseline (No Replay)} & 0 & 0 & 0.312 \\
\bottomrule
\end{tabular}
}
\caption{Computational overhead comparison. Our unified prompt achieves 1.29$\times$ speedup over sequential implementation with comparable PARG performance.}
\label{tab:replay_overhead}
\end{table}

\section{Additional Study of Personalized Editing}
\label{sec:editing_study}

We provide a detailed analysis of our personalized editing approach, addressing training data design.

\subsection{Training Data Design and Rationale}

\paragraph{Focus on Removal Instructions.}
Our training dataset primarily consists of concept removal instructions (e.g., ``Remove \texttt{<sks>} from the image''). While this may appear limited, we emphasize that \textbf{training data scope differs from evaluation scope}. The removal task serves as an effective training signal because:

\textbf{(1) Fine-Grained Localization:} Removal forces the model to precisely identify and locate the personalized concept, encoding fine-grained spatial and semantic information in learned concept tokens $\mathbf{P}^{(\text{gen})}$.

\textbf{(2) Leveraging Pretrained Capabilities:} By training only concept tokens while freezing all Transformer parameters (Section~\ref{subsec:architecture_clarification}), we specialize the concept representation for personalized editing while preserving the backbone's broad pretrained editing skills. This avoids catastrophic forgetting of diverse editing capabilities.

\textbf{(3) Data Efficiency:} Edit data can be straightforwardly constructed automatically via inpainting models, enabling scalable training data generation.

\subsection{Baseline Comparisons}

\paragraph{Editing Baselines.}
We acknowledge the possible concern regarding limited editing-specific baselines in Table~1. This reflects a fundamental challenge: to our knowledge, \textbf{OmniPersona is the first work to address instructional editing of newly learned personalized concepts} (represented as \texttt{<sks>} tokens).

\begin{figure*}[tb]

\centering
\includegraphics[width=1\linewidth]{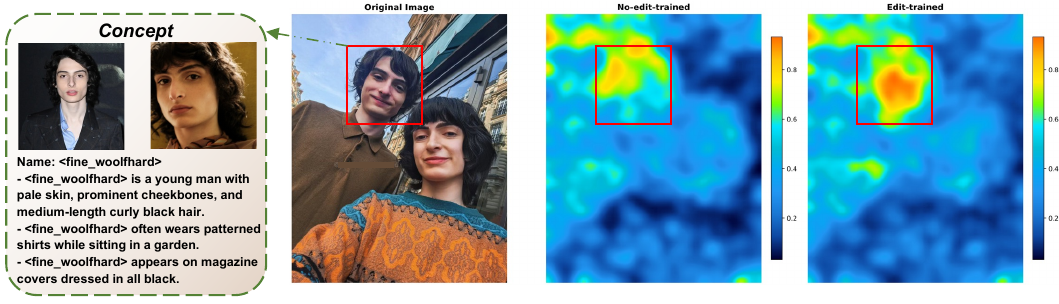}
\vspace{-7mm}
\captionof{figure}{Attention visualization showing focused localization of \texttt{<sks>} tokens after incorporating edit training data.
}

\vspace{-3mm}
\label{fig:attention}
\end{figure*}

\noindent\textbf{ Personalized Generation Methods (e.g., DreamBooth, IP-Adapter)}: These methods learn \texttt{<sks>} but focus on generation from scratch, not editing existing images. Adapting them for editing requires both \textbf{training an inversion model to encode input images} and \textbf{developing instruction-following mechanisms}.

This adaptation is non-trivial and constitutes a research contribution on its own, beyond the scope of comparing to OmniPersona.

\subsection{Detailed Analysis of Baseline Comparisons}

To provide a clearer illustration of the results presented in Table~1 of the main paper, we summarize the key baseline comparisons with detailed breakdowns.

\begin{table}[h]
\centering
\resizebox{0.95\linewidth}{!}{
\begin{tabular}{lcccl}
\toprule
\textbf{Method} & \textbf{SEMA-C} $\uparrow$ & \textbf{QUAL-I} $\uparrow$ & \textbf{Avg.} $\uparrow$ & \textbf{Personalization Strategy} \\
\midrule
Bagel+TP & 0.297 & 0.566 & 0.432 & Text prompt descriptions \\
GPT-4o+IP & 0.512 & 0.603 & 0.558 & Image prompts \\
\textbf{OmniPersona (Ours)} & \textbf{0.711} & \textbf{0.605} & \textbf{0.658} & Decoupled concept tokens \\
\bottomrule
\end{tabular}
}
\caption{Detailed breakdown of personalized editing performance across different personalization strategies. \textbf{Bagel+TP} uses the same backbone architecture but represents concepts via long-context text descriptions. \textbf{GPT-4o+IP} leverages GPT-4o's multimodal capabilities with few-shot image prompts.}
\label{tab:baseline_comparison}
\end{table}

\noindent\textbf{Few-Shot Prompting vs. Learned Tokens:} Bagel+TP achieves 0.432 average, demonstrating the limitations of text-only concept representations for fine-grained editing tasks. The 52.3\% performance gap indicates that learned concept tokens encode richer spatial and semantic information.
GPT-4o+IP achieves 0.558 average with notably lower semantic consistency (SEMA-C: 0.512) despite comparable image quality (QUAL-I: 0.603). This reveals challenges in preserving personalized identity through implicit few-shot learning.

The improvements stem from our decoupled token design mechanism, which enable specialized concept representations while leveraging pretrained capabilities.

\subsection{Attention Visualization}

We examined whether adding edit data during training would help, and visualized the attention between \texttt{<fine woolfhard>} and the image. Using challenging examples with the query "Can you point out where \texttt{<fine woolfhard>} is?", we observe that after incorporating edit data, the model's attention on \texttt{<fine woolfhard>} becomes more focused and captures the concept's fine-grained features more effectively.

\begin{figure}[htbp]

\centering
\includegraphics[width=1\linewidth]{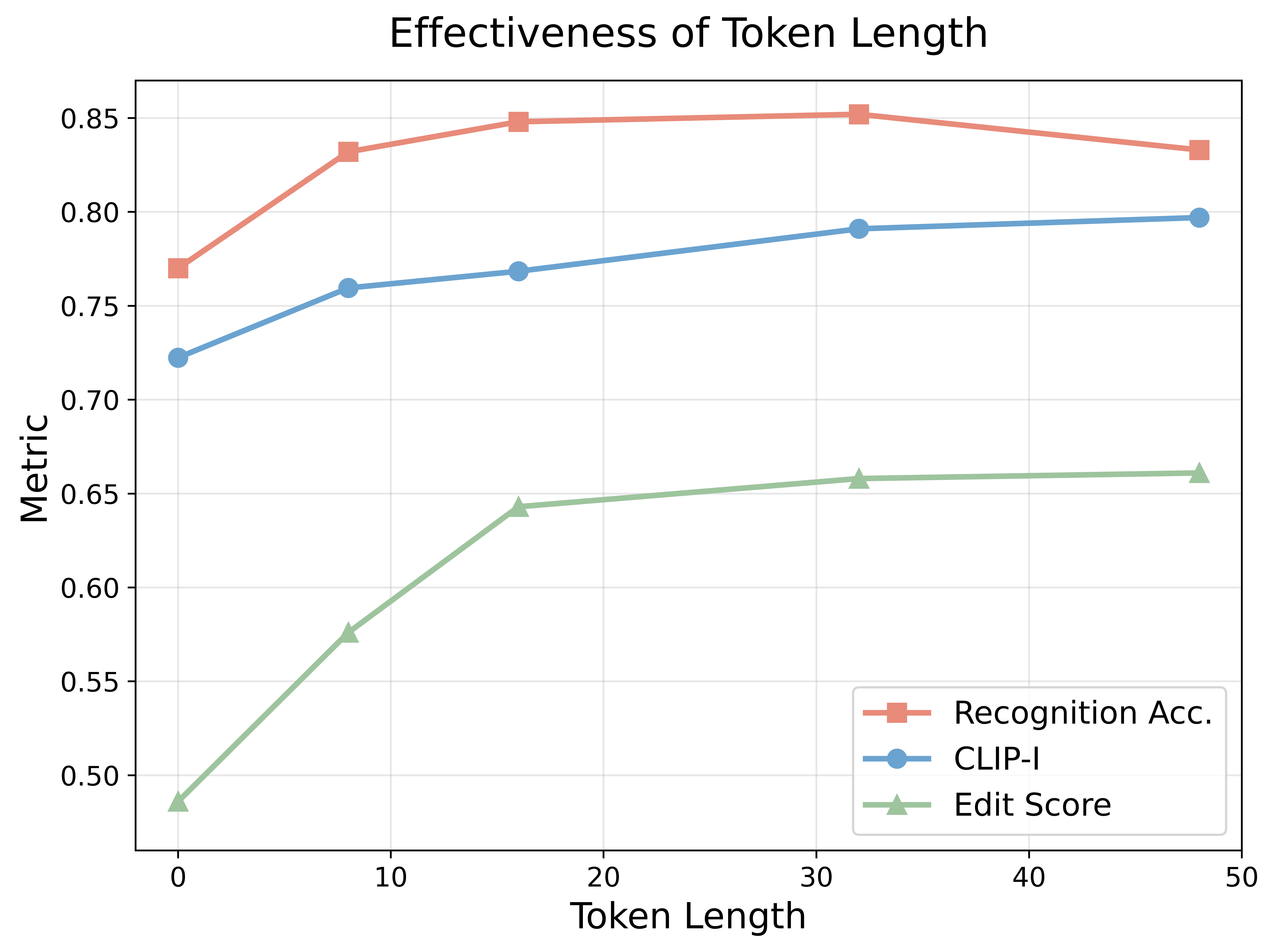}
\vspace{-7mm}
\captionof{figure}{Effect of learnable token number on model performance.
}

\vspace{-3mm}
\label{fig:token_ablation}
\end{figure}

\section{In-depth Study of Token Number}
\label{sec:token_number_study}

We conducted an ablation study on the number of learnable tokens and observed that performance improves as the number of tokens increases, but after reaching a certain point, the performance begins to decline. This phenomenon can be attributed to several factors.

Initially, increasing token numbers enhances the model's capacity to encode fine-grained personalized attributes and spatial features. With more tokens, the model can capture richer semantic information about the concept. However, beyond an optimal point, the added tokens introduce optimization challenges: the increased parameter space becomes harder to optimize with limited training data (only a few reference images per concept), leading to overfitting and degraded generalization.

\begin{table}[t]
\centering
\small
\begin{tabular}{ll}
\toprule
\textbf{Hyperparameter} & \textbf{Value} \\
\midrule
Optimizer & AdamW \\
Learning rate & $1 \times 10^{-4}$ with warmup \\
Training steps & 2,000 step per concept \\
Batch size & 8 \\
Weight decay & $0$ \\
Gradient clipping & Max norm 1.0 \\
Mixed precision & bfloat16 \\
\bottomrule
\end{tabular}
\caption{Training hyperparameters.}
\label{tab:train_hyperparameters}
\end{table}

\noindent\textbf{Information Redundancy.} Excessive tokens may encode redundant information, where multiple tokens learn similar features. This redundancy not only wastes model capacity but also dilutes the discriminative power of individual tokens, making it harder for the model to selectively attend to relevant concept features during generation and editing.

Based on our ablation results, we select 32 tokens (16 for understanding expert and 16 for generation expert) as the optimal configuration, balancing representation capacity and training efficiency.

\section{Additional Implement Detail}
\label{sec:implementation_details}
\subsection{Architecture Clarification}
\label{subsec:architecture_clarification}

We provide a detailed clarification on the architectural implementation of our Understanding Expert and Generation Expert Transformers, specifically addressing the parameter organization and routing mechanisms.

\paragraph{Complete Parameter Decoupling.}
The Understanding Expert and Generation Expert maintain \textbf{fully independent parameter sets}, not LoRA adapters or routing within a shared backbone. As defined in Eq.~(1), each expert branch $\mathcal{F}_{\text{und}}(\cdot)$ and $\mathcal{F}_{\text{gen}}(\cdot)$ comprises a complete Transformer with independent parameters:

\begin{itemize}
    \item \textbf{Understanding Expert:} $W^{(\text{und})}_{q,k,v}$, $\text{MLP}^{(\text{und})}$, $\text{LN}^{(\text{und})}_{\text{pre/post}}$
    \item \textbf{Generation Expert:} $W^{(\text{gen})}_{q,k,v}$, $\text{MLP}^{(\text{gen})}$, $\text{LN}^{(\text{gen})}_{\text{pre/post}}$
\end{itemize}

Crucially, \textbf{only the newly injected concept token embeddings} $\mathbf{P}^{(\text{und})}$ and $\mathbf{P}^{(\text{gen})}$ are trainable during personalization, while all expert-specific Transformer parameters ($W^{(\text{und})}$, $W^{(\text{gen})}$, MLPs, LayerNorms) are \textbf{frozen}. This parameter-efficient design enables few-shot concept learning without modifying the pretrained backbone. The forward pass through each expert follows:
\begin{equation}
\resizebox{0.9\linewidth}{!}{$
\mathbf{H}^{(\text{und})} =
\mathcal{F}_{\text{und}}\!\left(\mathbf{P}^{(\text{und})}, \mathbf{X}^{(\text{und})}\right), \quad
\mathbf{H}^{(\text{gen})} =
\mathcal{F}_{\text{gen}}\!\left(\mathbf{P}^{(\text{gen})}, \mathbf{X}^{(\text{gen})}\right),
$}
\end{equation}
where $\mathbf{P}^{(\text{und})} \in \mathbb{R}^{(N_u+1) \times d}$ and $\mathbf{P}^{(\text{gen})} \in \mathbb{R}^{N_g \times d}$ are the learnable concept token embeddings (initialized to zero), and $\mathbf{X}^{(\text{und})}$, $\mathbf{X}^{(\text{gen})}$ are the corresponding input embeddings routed to each expert.

\paragraph{Token Injection and Routing.}
Each concept is represented by 32 learnable embeddings structured as:
\[
\resizebox{0.9\linewidth}{!}{``$\textcolor{sksblue}{\texttt{<sks>}} \texttt{is}\, \textcolor{undgreen}{\texttt{<und\_1>}} \cdots \textcolor{undgreen}{\texttt{<und\_}N_u\texttt{>}} \; \textcolor{genpurple}{\texttt{<gen\_1>}} \cdots \textcolor{genpurple}{\texttt{<gen\_}N_g\texttt{>}}$.''}
\]
where $\texttt{<sks>}$ identifies the concept, $\{\texttt{<und\_}i\}$ are routed to the Understanding Expert, and $\{\texttt{<gen\_}j\}$ are routed to the Generation Expert. During forward propagation, each token is processed exclusively by its designated expert's parameters:
\begin{align}
\mathbf{Q, K, V}[i] &= W^{(\text{und})}_{q,k,v,o} \mathbf{X}[i], \quad \forall i \in \mathcal{I}_{\text{und}} \\
\mathbf{Q, K, V}[j] &= W^{(\text{gen})}_{q,k,v,o} \mathbf{X}[j], \quad \forall j \in \mathcal{I}_{\text{gen}}
\end{align}
This static, index-based routing ensures deterministic expert assignment based on token type.

\begin{table}[t]
\centering
\small
\begin{tabular}{l|ll}
\toprule
\textbf{Task} & \textbf{Hyperparameter} & \textbf{Value} \\
\midrule
\multirow{4}{*}{\textbf{Understanding}}
 & Max generation length & 1000 tokens \\
 & Sampling temperature & 0.3 \\
 & Do sample & False \\
 & Think mode & Disabled \\
\midrule
\multirow{10}{*}{\textbf{Generation}}
 & Image size & 512 $\times$ 512 \\
 & CFG text scale & 4.0 \\
 & CFG image scale & 1.0 \\
 & CFG interval & [0.4, 1.0] \\
 & Timestep shift & 3.0 \\
 & Diffusion steps & 50 \\
 & CFG renorm type & Global \\
 & CFG renorm min & 0.0 \\
 & Think mode & Disabled \\
\midrule
\multirow{10}{*}{\textbf{Editing}}
 & Image size & 512 $\times$ 512 \\
 & CFG text scale & 4.0 \\
 & CFG image scale & 2.0 \\
 & CFG interval & [0.0, 1.0] \\
 & Timestep shift & 3.0 \\
 & Diffusion steps & 50 \\
 & CFG renorm type & Text-channel \\
 & CFG renorm min & 0.0 \\
 & Think mode & Disabled \\
\bottomrule
\end{tabular}
\caption{Inference hyperparameters for different evaluation tasks.}
\label{tab:inference_hyperparameters}
\end{table}

\paragraph{Clarification on ``Shared Attention''.}
The term ``shared attention'' in Fig.~\ref{fig:method} refers to \textbf{computational flow sharing}, not parameter sharing. While $\mathbf{Q}$, $\mathbf{K}$, $\mathbf{V}$ are computed using expert-specific projections, all tokens participate in a single attention operation:
\begin{equation}
\mathbf{O} = \text{softmax}\left(\frac{\mathbf{Q}\mathbf{K}^\top}{\sqrt{d_k}}\right)\mathbf{V}
\end{equation}
This allows cross-expert information flow while maintaining parameter-level decoupling.

\subsection{Hyperparameter Configuration}
\label{subsec:hyperparameters}

We detail the hyperparameter settings used throughout our experiments, ensuring reproducibility.

\noindent\textbf{Training Hyperparameters.}
We optimize concept tokens using the AdamW optimizer.

\noindent\textbf{Inference Hyperparameters.}
We provide comprehensive inference configurations for understanding, generation, and editing tasks.

\noindent\textbf{Loss Function Weights.} 
As defined in Eq.~(9), we set equal weights: $\lambda_{\text{image}} = \lambda_{\text{edit}} = 400.0$.

\subsection{Evaluation Metrics}
\label{subsec:evaluation_metrics}

For personalized understanding, generation, and Attribute-Reasoning Generation, we follow the evaluation metrics provided by UnifyBench. For the personalized editing task, we modify the GEDIT evaluation template to conduct personalized assessment.

\paragraph{Personalized Understanding Metrics.}

\noindent\textbf{Recognition (Rec.):} We evaluate the model's ability to identify whether a given image contains the personalized concept. For each concept, we construct a balanced set of positive (containing the concept) and negative (not containing the concept) samples. Recognition accuracy is measured as:
\begin{equation}
\text{Rec.} = \frac{1}{2}\left(\text{Recall}_{\text{pos}} + \text{Recall}_{\text{neg}}\right)
\end{equation}
where $\text{Recall}_{\text{pos}}$ is the recall on positive samples and $\text{Recall}_{\text{neg}}$ is the recall on negative samples. This balanced metric ensures the model is not biased toward either predicting presence or absence.

\noindent\textbf{Visual Question Answering (VQA):} We assess the model's ability to answer visual questions about personalized concepts. We report two complementary metrics:
\begin{itemize}
    \item \textbf{VQA-BLEU:} BLEU score measuring n-gram overlap between predicted and reference answers, capturing lexical accuracy.
    \item \textbf{VQA-GPT:} GPT-4o-based evaluation scoring responses from 0 to 1 based on semantic alignment with key points in reference answers, capturing meaning beyond exact phrasing.
\end{itemize}

\noindent\textbf{Text-Only Question Answering (QA):} We evaluate the model's ability to answer text-only questions about learned concept attributes without visual input (e.g., ``Can you describe \texttt{<sks>}?''). This tests whether personalized knowledge is internalized in token representations. Metrics follow VQA: QA-BLEU and QA-GPT.

\paragraph{Personalized Generation Metrics.}

\begin{itemize}
\item \textbf{CLIP-I (Image Similarity):} We measure identity preservation by computing cosine similarity between CLIP image embeddings of generated images and reference images:
\begin{equation}
\resizebox{0.9\linewidth}{!}{$
\text{CLIP-I} = \frac{1}{N_{\text{gen}}N_{\text{ref}}} \sum_{i,j} \cos\left(\text{CLIP}_{\text{img}}(\mathbf{I}_{\text{gen}}^i), \text{CLIP}_{\text{img}}(\mathbf{I}_{\text{ref}}^j)\right)
$}
\end{equation}
Higher CLIP-I indicates better preservation of visual identity across generations.

\item \textbf{CLIP-T (Text-Image Alignment):} We measure prompt adherence by computing cosine similarity between CLIP text embeddings of generation prompts and CLIP image embeddings of generated images:
\begin{equation}
\resizebox{0.9\linewidth}{!}{$
\text{CLIP-T} = \frac{1}{N_{\text{gen}}} \sum_{i} \cos\left(\text{CLIP}_{\text{txt}}(\mathbf{T}_i), \text{CLIP}_{\text{img}}(\mathbf{I}_{\text{gen}}^i)\right)
$}
\end{equation}
This evaluates whether generated images faithfully reflect textual instructions.

\item \textbf{DINO (Perceptual Similarity):} We employ DINO-v2 features to measure semantic visual similarity between generated and reference images, providing a complementary signal to CLIP-I that captures object-level semantics:
\begin{equation}
\resizebox{0.9\linewidth}{!}{$
\text{DINO} = \frac{1}{N_{\text{gen}}N_{\text{ref}}} \sum_{i,j} \cos\left(\text{DINO}(\mathbf{I}_{\text{gen}}^i), \text{DINO}(\mathbf{I}_{\text{ref}}^j)\right)
$}
\end{equation}

\item \textbf{Face Similarity (Face-Simi):} For human subject concepts (10 out of 20 concepts in OmniPBench), we measure facial identity preservation using the pretrained ArcFace model:
\begin{equation}
\resizebox{0.9\linewidth}{!}{$
\text{Face-Simi} = \frac{1}{N_{\text{gen}}N_{\text{ref}}} \sum_{i,j} \cos\left(\text{ArcFace}(\mathbf{I}_{\text{gen}}^i), \text{ArcFace}(\mathbf{I}_{\text{ref}}^j)\right)
$}
\end{equation}
This metric is only computed for person concepts and provides a specialized measure of identity fidelity critical for personalized human generation.
\end{itemize}

\begin{figure}[htbp]

\centering
\includegraphics[width=1\linewidth]{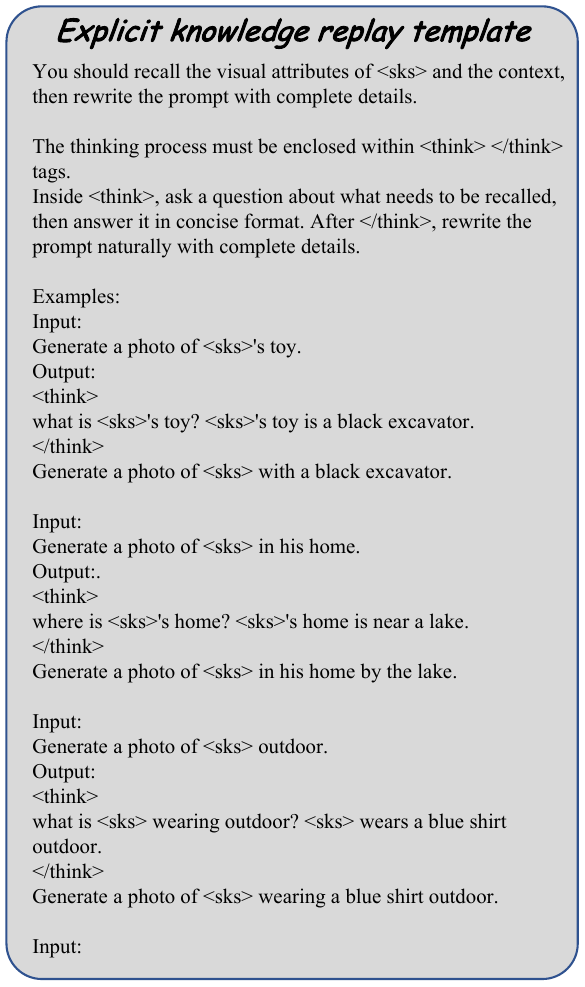}
\vspace{-7mm}
\captionof{figure}{Prompt template for explicit knowledge replay.
}

\vspace{-3mm}
\label{fig:replay_template}
\end{figure}

\begin{figure}[t]

\centering
\includegraphics[width=1\linewidth]{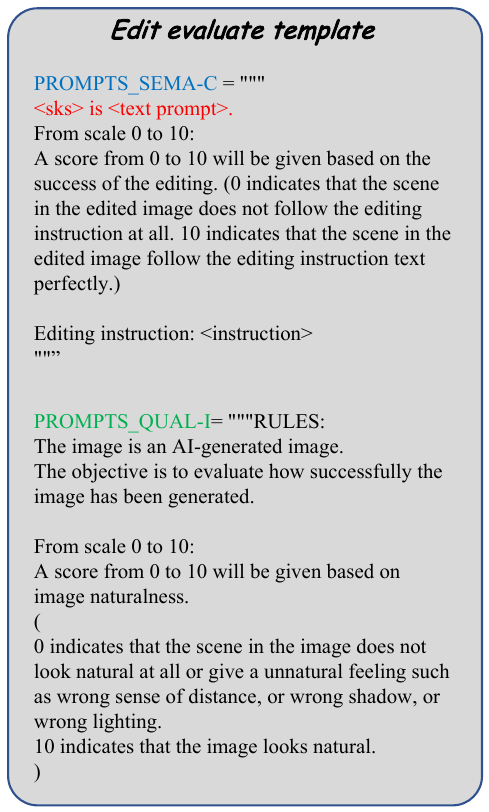}
\vspace{-7mm}
\captionof{figure}{Prompt template for personalized editing evaluation.
}

\vspace{-3mm}
\label{fig:edit_template}
\end{figure}

\paragraph{Personalized Attribute-Reasoning Generation (PARG) Metrics.}

PARG evaluates the model's ability to leverage \textit{textual} concept attributes during generation. For example, given ``Generate \texttt{<sks>} in his home'' without specifying home features, the model must recall learned attributes (e.g., ``\texttt{<sks>}'s home is by the sea''). We use:
\begin{itemize}
    \item \textbf{PARG-Score:} GPT-4o-based holistic evaluation (0-1 scale) assessing whether generated images correctly incorporate learned textual attributes.
    \item \textbf{PARG-CLIP-I:} CLIP-I computed between PARG generations and reference images, measuring identity preservation under attribute-conditioned generation.
\end{itemize}

\paragraph{Personalized Image Editing Metrics.}

\begin{itemize}
\item \textbf{Semantic Consistency (SEMA-C):} We employ GPT-4o as a judge to evaluate whether edited images faithfully follow editing instructions while preserving concept identity. The model is prompted to rate semantic alignment on a 0-1 scale based on instruction adherence.

\item \textbf{Quality of Image (QUAL-I):} We assess visual naturalness quality of edited images using GPT-4o-based evaluation (0-1 scale).
\end{itemize}

\noindent\textbf{Average Editing Score (Avg):} We report the arithmetic mean of SEMA-C and QUAL-I as the overall editing performance:
\begin{equation}
\text{Avg} = \frac{1}{2}\left(\text{SEMA-C} + \text{QUAL-I}\right)
\end{equation}

\section{Additional Qualitative Results}
\label{sec:rationale}
\begin{figure*}[tb]

\centering
\includegraphics[width=1\linewidth]{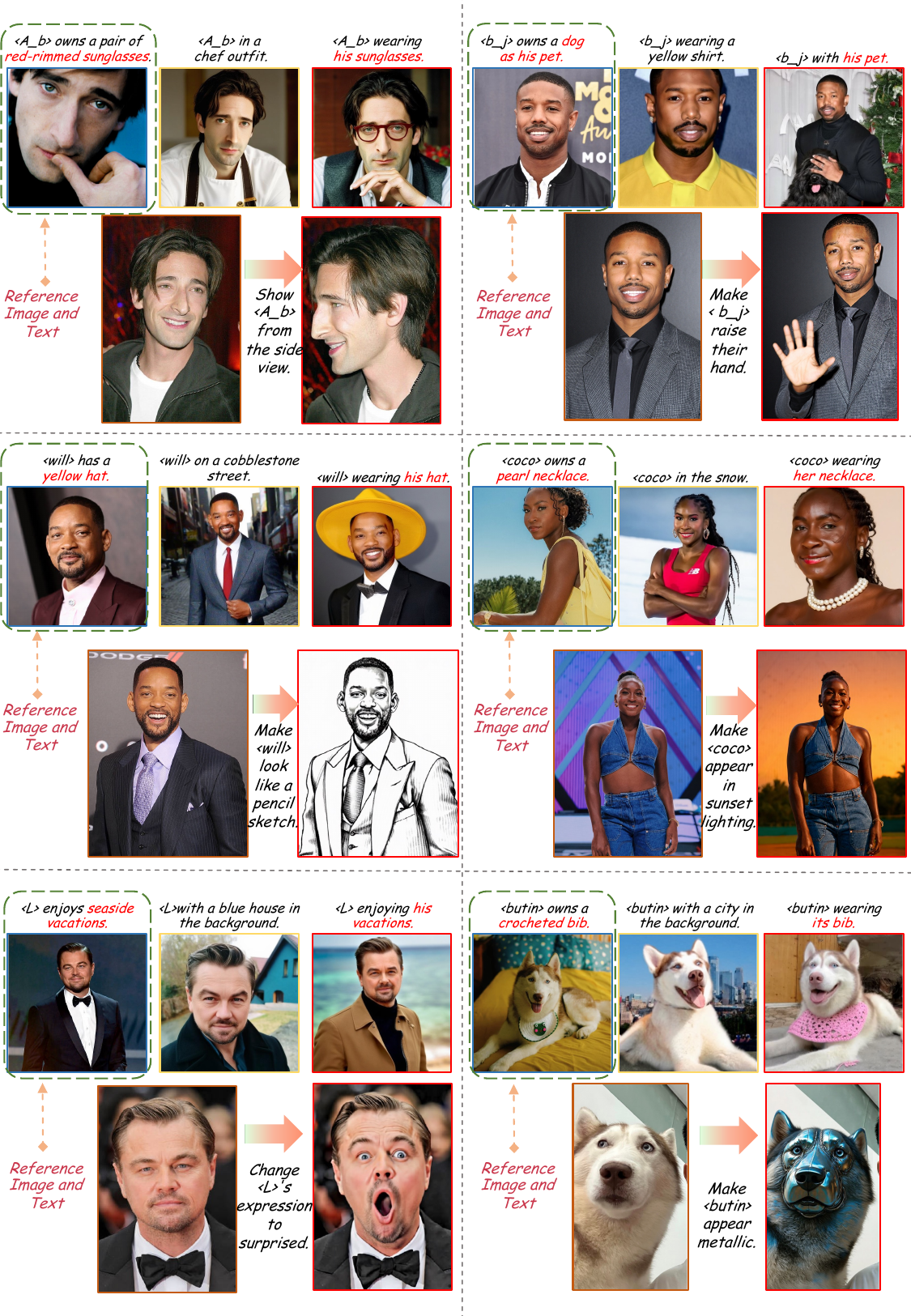}
\vspace{-7mm}
\captionof{figure}{Additional qualitative results for personalized generation and editing.
}

\vspace{-3mm}
\label{fig:more_case}
\end{figure*}

We provide more cases in Fig.~\ref{fig:more_case}.

\section{Limitations}
Due to limited personalized training data, we trained with only a small amount of edit data. As a result, editing in complex backgrounds remains challenging: the model may unintentionally alter backgrounds or fail to accurately localize the main subject. Although we used ablation studies to analyze attention maps with and without edit-data training, the improvements brought by edit data still require more comprehensive evaluation. In addition, while enhancing the similarity of the main subject, we observed a decline in instruction-following ability. This trade-off between subject preservation and instruction adherence in personalized settings requires further investigation.

\end{document}